\documentclass[sigconf]{acmart}
\usepackage{xspace}
\usepackage{amsmath}
\usepackage{multirow}
\usepackage{array}
\usepackage{booktabs}
\usepackage{cleveref}
\usepackage{balance}
\usepackage[linesnumbered, ruled, vlined]{algorithm2e}
\crefname{equation}{Eq.}{Eqs.}
\crefname{table}{Tab.}{Tables}
\crefname{figure}{Fig.}{Figures}
\crefname{algorithm}{Algorithm}{Algorithms}
\crefname{algocf}{Algorithm}{Algorithms}
\crefname{section}{Sec.}{Secs.}

\AtBeginDocument{%
  \providecommand\BibTeX{{%
    \normalfont B\kern-0.5em{\scshape i\kern-0.25em b}\kern-0.8em\TeX}}}

\setcopyright{acmcopyright}
\copyrightyear{2018}
\acmYear{2018}
\acmDOI{XXXXXXX.XXXXXXX}

\acmConference[Conference acronym 'XX]{Make sure to enter the correct
  conference title from your rights confirmation emai}{June 03--05,
  2018}{Woodstock, NY}
\acmPrice{15.00}
\acmISBN{978-1-4503-XXXX-X/18/06}

\copyrightyear{2023}
\acmYear{2023}
\setcopyright{acmlicensed}\acmConference[WWW '23]{Proceedings of the ACM Web Conference 2023}{May 1--5, 2023}{Austin, TX, USA}
\acmBooktitle{Proceedings of the ACM Web Conference 2023 (WWW '23), May 1--5, 2023, Austin, TX, USA}
\acmPrice{15.00}
\acmDOI{10.1145/3543507.3583353}
\acmISBN{978-1-4503-9416-1/23/04}
\begin{document}

\title{SINCERE: Sequential Interaction Networks representation learning on Co-Evolving RiEmannian manifolds}



\settopmatter{authorsperrow=3}
\author{Junda Ye}
\thanks{Contact Junda Ye (jundaye@bupt.edu.cn) and Li Sun (ccesunli@ncepu.edu.cn) for any question.}
\affiliation{%
  \institution{Beijing University of Posts and Telecommunications}
  \city{Beijing}
  \country{China}
}
\email{jundaye@bupt.edu.cn}

\author{Zhongbao Zhang}
\affiliation{%
  \institution{Beijing University of Posts and Telecommunications}
  \city{Beijing}
  \country{China}
}
\email{zhongbaozb@bupt.edu.cn}

\author{Li Sun}
\affiliation{%
  \institution{North China Electric Power University}
  \city{Beijing}
  \country{China}
}
\email{ccesunli@ncepu.edu.cn}

\author{Yang Yan}
\affiliation{%
  \institution{Beijing University of Posts and Telecommunications}
  \city{Beijing}
  \country{China}
}
\email{yanyang42@bupt.edu.cn}

\author{Feiyang Wang}
\affiliation{%
  \institution{Beijing University of Posts and Telecommunications}
  \city{Beijing}
  \country{China}
}
\email{fywang@bupt.edu.cn}

\author{Fuxin Ren}
\affiliation{%
  \institution{Beijing University of Posts and Telecommunications}
  \city{Beijing}
  \country{China}
}
\email{renfuxin@bupt.edu.cn}


\newcommand{\ourmethod}{SINCERE\xspace}
\newcommand{\eg}{\textit{e.g.}}
\newcommand{\ie}{\textit{i.e.}}
\newcommand{\etc}{\textit{etc.}}
\newcommand{\etal}{\textit{{\sl et al.\ }}}
\newcommand{\wrt}{\textit{{\sl w.r.t. }}}
\newcommand{\operator}{\textsc{}}
\renewcommand{\shortauthors}{Junda Ye, \etal}

\newcommand{\blue}[1]{{{\textcolor{blue}{#1}}}}
\newcommand{\blueno}[1]{{{\textcolor{blue}{#1}}}}
\newcommand{\red}[1]{{{\textcolor{red}{#1}}}}
\newcommand{\redno}[1]{{{\textcolor{red}{#1}}}}
\newcommand{\orange}[1]{{{\textcolor{orange}{#1}}}}
\newcommand{\orangeno}[1]{{{\textcolor{orange}{#1}}}}
\newcommand{\purple}[1]{{{\textcolor{Magenta}{#1}}}}
\newcommand{\purpleno}[1]{{{\textcolor{Magenta}{#1}}}}
\newcommand{\green}[1]{{{\textcolor{green}{#1}}}}
\newcommand{\greenno}[1]{{{\textcolor{green}{#1}}}}
\newcommand{\cyan}[1]{{{\textcolor{cyan}{#1}}}}
\newcommand{\yellow}[1]{{{\textcolor{yellow}{#1}}}}
\newcommand{\reminder}[1]{{{\textcolor{green}{#1}}}}
\newcommand{\fuchsia}[1]{{{\textcolor{fuchsia}{#1}}}}
\newcommand{\fuchsiano}[1]{{{\textcolor{fuchsiano}{#1}}}}
\newcommand{\black}[1]{{{\textcolor{black}{#1}}}}
\newcommand{\blackno}[1]{{{\textcolor{black}{#1}}}}
\newcommand{\sketchgreen}[1]{{{\textcolor{sketchgreen}{#1}}}}
\definecolor{expblue}{RGB}{84,113,171}
\definecolor{expred}{RGB}{182,86,85}
\definecolor{exporange}{RGB}{209,136,92}
\definecolor{expgreen}{RGB}{106,166,110}
\definecolor{sketchgreen}{RGB}{71, 136, 50}

\begin{abstract}
Sequential interaction networks (SIN) have been commonly adopted in many applications such as recommendation systems, search engines and social networks to describe the mutual influence between users and items/products. Efforts on representing SIN are mainly focused on capturing the dynamics of networks in Euclidean space, and recently plenty of work has extended to hyperbolic geometry for implicit hierarchical learning. Previous approaches which learn the embedding trajectories of users and items achieve promising results. However, there are still a range of fundamental issues remaining open. For example, is it appropriate to place user and item nodes in one identical space regardless of their inherent discrepancy? Instead of residing in a single fixed curvature space, how will the representation spaces evolve when new interaction occurs? 

To explore these implication for sequential interaction networks, we propose \textbf{\ourmethod}, a novel method representing \textbf{\underline{S}}equential \textbf{\underline{I}}nteraction \textbf{\underline{N}}etworks on \textbf{\underline{C}}o-\textbf{\underline{E}}volving \textbf{\underline{R}}i\textbf{\underline{E}}mannian manifolds. \ourmethod not only takes the user and item embedding trajectories in respective spaces into account, but also emphasizes on the space evolvement that how curvature changes over time. Specifically, we introduce a fresh cross-geometry aggregation which allows us to propagate information across different Riemannian manifolds without breaking conformal invariance, and a curvature estimator which is delicately designed to predict global curvatures effectively according to current local Ricci curvatures. Extensive experiments on several real-world datasets demonstrate the promising performance of \ourmethod over the state-of-the-art sequential interaction prediction methods.
\end{abstract}




\maketitle

\section{Introduction}
Predicting future interactions for sequential interaction networks is essential in many domains, such as 
``Guess You Like'' in e-commerce recommendation systems \citep{he2014predicting, jiang2016learning}, ``Related Searches'' in search engines \citep{latentcross, huang2018entity} and ``Suggested Posts'' in social networks \citep{HTGN, CAW}. Specifically, purchasers and commodities form a transaction interaction network recording the trading or rating behaviors on e-commerce platforms, predicting interactions accurately helps improve the quality of item recommendations. Users and posts constitute a social interaction network indicating whether and when a user will click/comment posts, early interventions can be made to prevent teenagers from malicious posts like Internet fraud on social media with the help of interaction prediction.

Graph representation learning plays an important role in demonstrating different nodes in graphs within low-dimensional embeddings. In the literature, the majority of existing studies choose Euclidean space as representation space for its simplicity and efficiency. However, for certain graph topology patterns (\eg, hierarchical, scale-free or spherical data), Euclidean space may suffer from large distortion \citep{chami2019hyperbolic}. Recently, non-Euclidean spaces have emerged as an alternative for their strong capability to embed hierarchical or cyclical data into a low-dimensional space while preserving the structure \citep{mathieu2019continuous, HAN, ZhangWSLS21}. Specifically, hyperbolic geometry was first introduced to graph representation learning \citep{nickel2017poincare, nickel2018learning, HNN}, and 
\begin{table}[ht]
    \caption{The average $\delta$-hyperbolicity \citep{Gromov1987} of user and item subgraphs on Mooc and Wikipedia dataset along the timeline.}
    \setlength{\tabcolsep}{12pt}{
    \begin{tabular}{l | c c c}
        \toprule
        \specialrule{0em}{1pt}{1pt}
        \textbf{Dataset} & \textbf{Timeline} & \textbf{User} & \textbf{Item} \\
        \hline
        \multirow{3}{*}{Mooc} & start & 0.77 & 0.50 \\
         & middle & 1.0 & 0.67 \\
         & end & 1.0 & 0.13 \\
        \hline
        \multirow{3}{*}{Wikipedia} & start & 1.04 & 1.16 \\
         & middle & 1.10 & 1.30 \\
         & end & 1.33 & 1.40 \\ 
        \specialrule{0em}{1pt}{0pt}
        \bottomrule
    \end{tabular}}
    \label{study case table}
\end{table}
\begin{figure}[ht]
    \centering
    \includegraphics[width=0.95\linewidth]{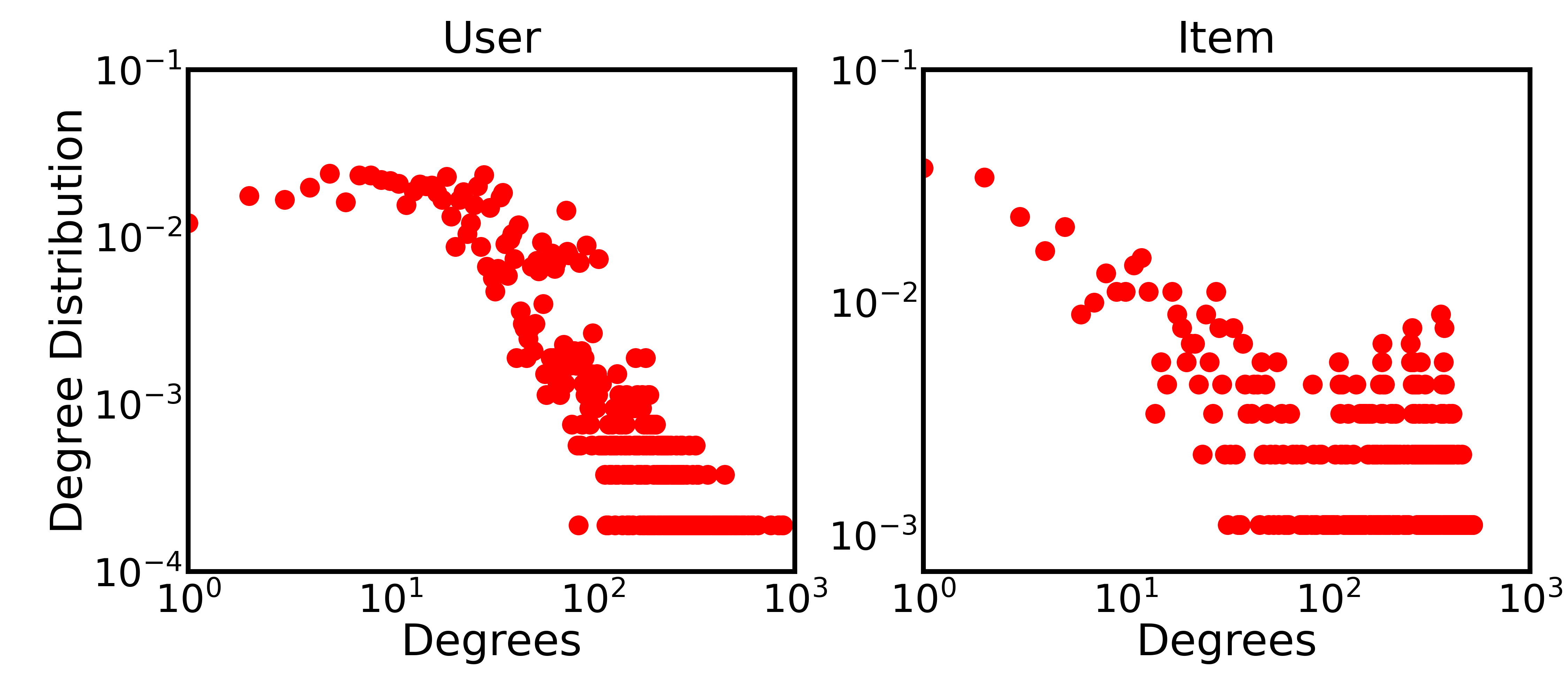}
    \vspace{-2mm}
    \caption{User and item degree distribution on Wikipedia.}
    \label{study case figure}
    \Description{Study case of user and item degree distribution on Wikipedia.}
\end{figure}
then spherical spaces have also gained much attention \citep{defferrard2020deepsphere, rezende2020normalizing}. The development of representation spaces is prosperous nowadays. 


With regard to representing sequential interaction networks, a significant amount of work has been proposed to learn the user and item embeddings from their historical interaction records in Euclidean space \citep{tgsrec, BiGI, deepcoevolve, CTDNE, deepred, latentcross}. Recurrent models such as Time-LSTM \citep{time-lstm}, Time-Aware LSTM \citep{tlstm} and RRN \citep{RRN} capture dynamics of users and items by endowing them with a long short-term memory. HILI \citep{hili} is the successor of Jodie \citep{jodie} and both of them are promising methods for real-world temporal interaction graphs. CTDNE \citep{CTDNE} adopts random walk on temporal networks and CAW \citep{CAW} makes it causal and anonymous to inductively represent sequential networks. Additionally, recent models have been proposed to embed sequential interaction networks into hyperbolic spaces \citep{chamberlain2019scalable, HME, HGCF, mirvakhabova2020performance, dkabrowski2021efficient}. HyperML \citep{hyperml} investigates the notion of learning user and item representations in terms of metric learning in hyperbolic space. HTGN \citep{HTGN} designs a recurrent architecture on the sequence of snapshots under hyperbolic geometric. 


Existing interaction network representation methods did have achieved promising results. However, they still suffer from two major shortcomings. 
\textbf{First}, they simply place \textit{two different} types of nodes (user and item nodes) in \textit{one identical} space, which is ambiguous and counter-intuitive. For instance, viewers and films are two kinds of nodes with totally different characters and distributions which are usually described by the notion of curvature in Riemannian geometry \citep{wilson2014spherical, sreejith2016forman, bachmann20a}. Here, we give motivated examples in \cref{study case figure} and \cref{study case table} which show that not only the structure of user and item networks varies according to time, but also the degree distribution of users and items are different from each other. Therefore, setting user and item nodes in two spaces seems a properer way and designing an effective message passing mechanism considering cross-space scenarios is challenging. 
\textbf{Second}, they assume the space is static but in fact the interaction network constantly \textit{evolves} over time, which is mainly ignored in the previous studies. More specifically, they attempt to learn the representations on a single fixed curvature manifold, either Euclidean space or the standard Poincar{\'{e}} Ball. The static curvature is a strong prerequisite that cannot manifest the representation space of inherent dynamics when new edges of the interaction networks are continuously formed. The challenge here is how to estimate the global curvatures accurately for the next period from local structure information. 
Besides, most of the existing methods \citep{jodie, hili, time-lstm, tlstm} also assume that an interaction would only influence corresponding nodes while ignoring the information propagation among the same type of nodes implicitly. New methods are also needed to incorporate the information flow between the same type of nodes.

To overcome the aforementioned shortcomings and challenges, we propose \textbf{\ourmethod} to represent \textbf{\underline{S}}equential \textbf{\underline{I}}nteraction \textbf{\underline{N}}etworks on \textbf{\underline{C}}o-\textbf{\underline{E}}volving \textbf{\underline{R}}i\textbf{\underline{E}}mannian manifolds. 
Specifically, \ourmethod sets both user and item nodes in two $\kappa$-stereographic spaces, interactions with time embeddings in Euclidean space simultaneously. Next, it estimates the sectional curvatures of two co-evolving spaces according to local structure information through a curvature neural network with shared parameters. Then, the dissemination and integration of information, as well as node embeddings updating are performed through cross-geometry aggregation and updating components in our model. Finally, \ourmethod is mainly learned through a Pull-and-Push loss based on maximum likelihood optimization. Extensive experiments validate the performance of \ourmethod on several datasets.

The major contributions of this paper are summarized as follows:
\begin{itemize}
    \item In this paper, we uncover the intrinsic differences between user and item nodes in sequential interaction networks, and model the dynamicity of representation spaces. It is the first attempt to represent sequential interaction networks on a pair of co-evolving Riemannian manifolds to the best we know.
    \item We propose a novel model \ourmethod for sequential interaction networks representation learning. 
    \ourmethod represents user and item nodes in dual $\kappa$-stereographic spaces with variable curvatures, encoding both immanent dissimilarities and temporal effects of user and item nodes. 
    \item We conduct extensive experiments to evaluate our model with multiple baselines for the tasks of interaction prediction. Experimental results show the superiority and effectiveness of our proposed model.
\end{itemize}

\section{Preliminaries}
\subsection{Riemannian Manifold}
A Riemannian manifold $(\mathcal{M}, g)$ is a real and smooth manifold equipped with an inner product on tangent space $g_{\mathbf{x}}: \mathcal{T}_{\mathbf{x}}\mathcal{M} \times \mathcal{T}_{\mathbf{x}}\mathcal{M} \rightarrow \mathbb{R}$ at each point $\mathbf{x} \in \mathcal{M}$, where the tangent space $\mathcal{T}_{\mathbf{x}}\mathcal{M}$ is the first order approximation of $\mathcal{M}$ around $\mathbf{x}$, which can be locally seemed Euclidean. The inner product metric tensor $g_{\mathbf{x}}$ is both positive-definite and symmetric, \ie, $g_{\mathbf{x}}(\mathbf{v}, \mathbf{v}) \geq 0, g_{\mathbf{x}}(\mathbf{u}, \mathbf{v}) = g_{\mathbf{x}}(\mathbf{v}, \mathbf{u})$ where $\mathbf{x} \in \mathcal{M}$ and $ \mathbf{u}, \mathbf{v} \in \mathcal{T}_{\mathbf{x}}\mathcal{M}$.


\subsection{The \texorpdfstring{$\kappa-$}.stereographic Model}
Hyperboloid and sphere are two common models for hyperbolic and spherical Riemannian geometry. However, hyperbolic space is not a typical vector space, which means we cannot calculate the embedding vectors with operators as if we use in Euclidean space. In order to overcome the gap among hyperbolic, spherical and euclidean geometry, we adopt the $\kappa$-stereographic model \citep{bachmann20a} as it unifies operations with gyrovector formalism \citep{ungar2008gyrovector}. 

An $n$-dimension $\kappa$-stereographic model is a smooth manifold $\mathcal{M}_{\kappa}^{n}=\left\{\mathbf{x} \in \mathbb{R}^{n} \mid-\kappa\|\mathbf{x}\|_{2}^{2}<1\right\}$ equipped with a Riemannian metric $g_{\mathbf{x}}^{\kappa}=\left(\lambda_{\mathbf{x}}^{\kappa}\right)^{2} \mathbf{I}$, where $\kappa \in \mathbb{R}$ is the sectional curvature and $\lambda_{\mathbf{x}}^{\kappa}=2\left(1+\kappa\|\mathbf{x}\|_2^{2}\right)^{-1}$ is the point-wise conformal factor. To be specific, $\mathcal{M}_{\kappa}^{n}$ is the stereographic projection model for spherical space ($\kappa > 0$) and the Poincar{\'{e}} ball model of radius $1/\sqrt{-\kappa}$ for hyperbolic space ($\kappa < 0$), respectively. We first list requisite operators below, specific ones are attached in Appendix. Note that bold letters denote the vectors on the manifold.

\textbf{M$\boldsymbol{\ddot{o}}$bius Addition.} It is a non-associative addition $\oplus_{\kappa}$ of two points $\mathbf{x}, \mathbf{y} \in \mathcal{M}_{\kappa}^{n}$ defined as:
\begin{equation} \label{mobius addition}
    \begin{aligned}
        \mathbf{x} \oplus_{\kappa} \mathbf{y} &=\frac{\left(1-2 \kappa\langle\mathbf{x}, \mathbf{y}\rangle-\kappa\|\mathbf{y}\|_{2}^{2}\right) \mathbf{x}+\left(1+\kappa\|\mathbf{x}\|_{2}^{2}\right) \mathbf{y}}{1-2 \kappa\langle\mathbf{x}, \mathbf{y}\rangle+\kappa^{2}\|\mathbf{x}\|_{2}^{2}\|\mathbf{y}\|_{2}^{2}}.
    \end{aligned}
\end{equation}

\textbf{M$\boldsymbol{\ddot{o}}$bius Scalar and Matrix Multiplication.} The augmented M$\ddot{o}$bius scalar and matrix multiplication $\otimes_{\kappa}$ of $\mathbf{x} \in \mathcal{M}_{\kappa}^{n} \backslash \left\{\mathbf{o}\right\}$ by $r \in \mathbb{R}$ and $\boldsymbol{M} \in \mathbb{R}^{m \times n}$ is defined as follows:
\begin{align}
    r \otimes_{\kappa} \mathbf{x} = &\frac{1}{\sqrt{\kappa}} \tanh \left(r \tanh ^{-1}(\sqrt{\kappa}\|\mathbf{x}\|_{2})\right) \frac{\mathbf{x}}{\|\mathbf{x}\|_{2}}, \label{mobius scalar multiplication} \\
    \boldsymbol{M} \otimes_{\kappa} \mathbf{x} =(1 / \sqrt{\kappa}) &\tanh \left(\frac{\|\boldsymbol{M} \mathbf{x}\|_2}{\|\mathbf{x}\|_2} \tanh ^{-1}(\sqrt{\kappa}\|\mathbf{x}\|_2)\right) \frac{\boldsymbol{M} \mathbf{x}}{\|\boldsymbol{M} \mathbf{x}\|_2}. \label{mobius matrix multiplication}
\end{align}

\textbf{Exponential and Logarithmic Maps.} The connection between the manifold $\mathcal{M}_{\kappa}^{n}$ and its point-wise tangent space  $\mathcal{T}_{\mathbf{x}}\mathcal{M}_{\kappa}^{n}$ is established by the exponential map $\exp_{\mathbf{x}}^{\kappa}: \mathcal{T}_{\mathbf{x}}\mathcal{M}_{\kappa}^{n} \rightarrow \mathcal{M}_{\kappa}^{n}$ and logarithmic map $\log_{\mathbf{x}}^{\kappa}: \mathcal{M}_{\kappa}^{n} \rightarrow \mathcal{T}_{\mathbf{x}}\mathcal{M}_{\kappa}^{n}$ as follows:
\begin{align}
    \exp _{\mathbf{x}}^{\kappa}(\mathbf{v}) =\mathbf{x} \oplus_{\kappa}&\left(\tan _{\kappa}\left(\sqrt{|\kappa|} \frac{\lambda_{\mathbf{x}}^{\kappa}\|\mathbf{v}\|_{2}}{2}\right) \frac{\mathbf{v}}{\|\mathbf{v}\|_{2}}\right), \label{exp map} \\
    \log _{\mathbf{x}}^{\kappa}(\mathbf{y}) =\frac{2}{\lambda_{\mathbf{x}}^{\kappa}\sqrt{|\kappa|}} &\tan _{\kappa}^{-1}\left\|-\mathbf{x} \oplus_{\kappa} \mathbf{y}\right\|_{2} \frac{-\mathbf{x} \oplus_{\kappa} \mathbf{y}}{\left\|-\mathbf{x} \oplus_{k} \mathbf{y}\right\|_{2}}. \label{log map}
\end{align}

\textbf{Distance Metric.} The generalized distance $d_{\mathcal{M}}^{\kappa}\left(\cdot, \cdot\right)$ between any two points $\mathbf{x}, \mathbf{y} \in \mathcal{M}_{\kappa}^{n}, \mathbf{x} \neq \mathbf{y}$ is defined as:
\begin{equation} \label{distance metric}
    \begin{aligned}
        d_{\mathcal{M}}^{\kappa}\left(\mathbf{x}, \mathbf{y}\right) = \frac{2}{\sqrt{|\kappa|}}\tan_{\kappa}^{-1}\left(\sqrt{|\kappa|} \left\|-\mathbf{x} \oplus_{k} \mathbf{y}\right\|_{2}\right).
    \end{aligned}
\end{equation}

\begin{figure*}[ht]
    \centering
    \includegraphics[width=0.91\linewidth]{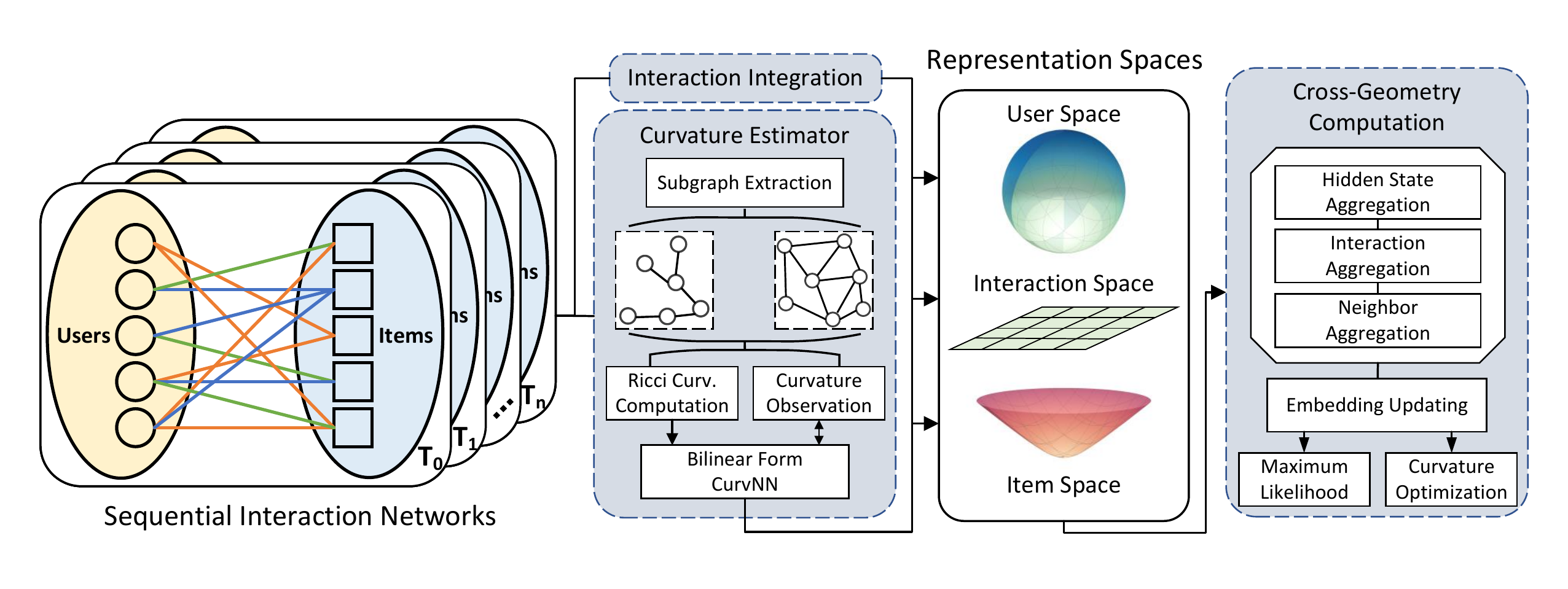}
    \caption{Illustration of \ourmethod architecture. We first divide timestamps into several intervals, each of which can be regarded as a batch. Interactions with different colors refer to different timestamps. Next, the graph is sent to the interaction integration component and the curvature estimator for capturing time and structure information, respectively. Then, user and item nodes in different Riemannian manifolds pass neighbor information to each other through novel cross-geometry aggregation and updating. Last, the representation and parameters of \ourmethod are optimized through maximum likelihood training objective. }
    \label{architecture}
    \Description{Architecture of our method.}
\end{figure*}

\subsection{Sectional Curvature and Ricci Curvature}
In Riemannian geometry, curvature is the amount by which a curve deviates from being a straight line, or a surface deviates from being a plane, which is originally defined on the continuous smooth manifold.
Sectional curvature and Ricci curvature are two success formulations for the discrete graph \citep{Ye2020Curvature}, describing the global and local structure, respectively.

\textbf{Sectional Curvature.} Sectional curvature is an equivalent but more geometrical description of the curvature of Riemannian manifolds compared to the Riemann curvature tensor \citep{lee2018introduction}. It is defined over all two-dimensional subspaces passing through a point and is treated as the constant curvature of a Riemannian manifold in recent work.

\textbf{Ricci Curvature.} Ricci curvature is the average of sectional curvatures and is mathematically a linear operator on tangent space at a point. In graph domain, it is extended to measure how the geometry of a pair of neighbors deviates from the case of gird graphs locally. Ollivier-Ricci \citep{ollivier2009ricci} is a coarse approach to computing Ricci curvature while Forman-Ricci \citep{forman2003bochner} is combinatorial and faster to compute.

\begin{table}[ht] 
    \caption{Table of symbols used in the paper} \label{tab:symbols}
    \centering
    \begin{tabular}{c | l}
         \toprule
         Symbol & Description \\
         \midrule
         $\mathbb{U}_{\kappa_u}$ & User space with curvature $\kappa_{u}$ \\
         $\mathbb{I}_{\kappa_i}$ & Item space with curvature $\kappa_{i}$ \\
         $T_n$ & Time interval from $t_{n-1}$ to $t_n$ \\
         $\kappa_{u/i}^{T}$ & Curvature of user/item space during $T$ \\
         $\mathbf{u}(T)$ & Embedding of user $u$ during $T$ \\ 
         $\mathbf{i}(T)$ & Embedding of item $i$ during $T$ \\
         $\mathcal{N}_{u/i}^{T}$ & Neighbors of user/item during $T$ \\
         $\mathcal{E}_{u/i}^{T}$ & Edges linked to user/item during $T$ \\
         \bottomrule
    \end{tabular}
\end{table}

\vspace{-2mm}
\subsection{Problem Definition and Notation}
Before presenting our proposed model, we summarize the important notations in \cref{tab:symbols}. We may drop the subscript indices in later sections for clarity. The definitions of sequential interaction network and representation learning task are stated as follows:

\newtheorem*{def1}{Definition 1 (Sequential Interaction Network (SIN))} 
\begin{def1}
Let $\mathcal{G} = \{\mathcal{U, I, E, T, X}\}$ denote a sequential interaction network, where $\mathcal{U}$ and $\mathcal{I}$ are two disjoint node sets of user and item, respectively. $\mathcal{E} \subseteq \mathcal{U} \times \mathcal{I} \times \mathcal{T}$ denotes the interaction/edge set, and $\mathcal{X} \in \mathbb{R}^{|\mathcal{E}| \times d_{\mathcal{X}}}$ is the attributes associated with the interaction. $d_{\mathcal{X}}$ represents feature dimension. Each edge $e \in \mathcal{E}$ can be indicated as a triplet $e = (u, i, t)$ where $u \in \mathcal{U}, i \in \mathcal{I}$, and $t \in \mathcal{T}$ is the timestamp recording that user $u$ and item $i$ interact with each other at time $t$.
\end{def1}

\newtheorem*{def2}{Definition 2 (SIN Representation Learning Task)} 
\begin{def2}
For a sequential interaction network $\mathcal{G}$, it aims to learn encoding functions $\Phi_u: \mathcal{U} \rightarrow \mathbb{U}_{\kappa_u}$, $\Phi_i: \mathcal{I} \rightarrow \mathbb{I}_{\kappa_i}$ and $\Phi_e: \mathcal{E} \rightarrow \mathbb{E}$ that map two kinds of nodes in networks and interaction between them into dynamic low-dimensional embedding spaces under different Riemannian geometry, where nodes and interaction are represented as compact embedding vectors. Meanwhile, the embeddings capture the intrinsic relation between two types of nodes and are capable of predicting future interactions effectively.
\end{def2}

\section{SINCERE}
Recall that the motivated examples given previously show the intrinsic difference between users and items, to solve the main issues of existing methods on representing sequential interaction networks, we consider the embeddings of user and item residing in different Riemannian manifolds. The two Riemannian manifolds are then bridged by a Euclidean tangent space, which is the representation space of interactions. How the representation spaces evolve over time is told by a curvature estimator.

The overall framework of SINCERE is presented in \cref{architecture}. The components of our approach are introduced in the following.

\vspace{-2mm}
\subsection{User-Item Co-Evolving Mechanism} 
\label{geometry cross}
\begin{figure}[ht]
    \centering
    \includegraphics[width=\linewidth]{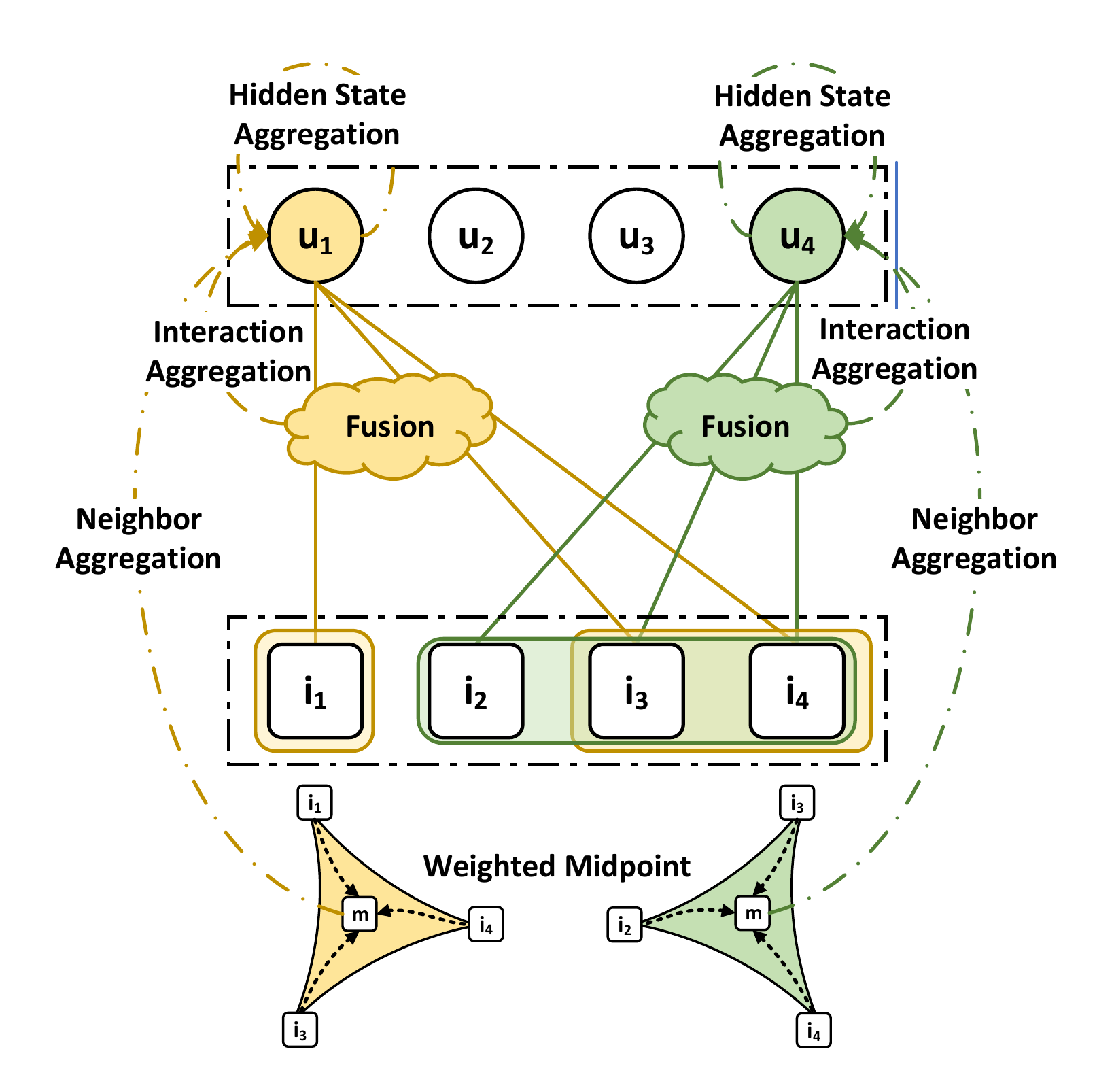}
    \caption{Illustration of one step aggregation. An example of how users update their embeddings from: i) hidden representations, ii) fusion of associated interactions, and iii) weighted midpoint of their neighbors.}
    \label{one step aggregation}
    \Description{Illustration of one step aggregation.}
\end{figure}
As we place the user and item nodes on different Riemannian manifolds, they are recorded by the embedding matrix $\mathbf{U}$ and $\mathbf{I}$ respectively, which can be initialized with the feature vectors such as the user profiling or film description if available. Since the initialized $\mathbf{U}$ and $\mathbf{I}$ are in Euclidean space, we logarithmically map them to corresponding Riemannian space $\mathbb{U}_{\kappa_u}$ and $\mathbb{I}_{\kappa_i}$ with \cref{log map}.


For the message propagation and information aggregation, 
the design of \ourmethod not only incorporates the information across different geometric spaces via a unified formalism, but also guarantees that all operations are conformal invariant. For clarity of expressions, we define $\operatorname{map}$ operation to map the vectors in manifold $\mathcal{M}_{\kappa_1}$ to $\mathcal{M}_{\kappa_2}$ as $\operatorname{map}_{\kappa_1}^{\kappa_2}\left(\cdot\right):=\exp_{\mathbf{o}}^{\kappa_2}\left(\log_{\mathbf{o}}^{\kappa_1}\left(\cdot \right)\right)$. The aggregation operators of user and item are defined as follows:
\begin{equation}
\begin{split}
    \mathbf{u}_{j}(T_n) \leftarrow \underbrace{\boldsymbol{M}_1 \otimes_{\kappa_u^T} \mathbf{h}_{u_j}(T_n)}_{\textit{user hidden aggregation}} 
    & \oplus_{\kappa_u^T}  \underbrace{\boldsymbol{M}_2 \otimes_{\kappa_u^T} \exp_{\mathbf{o}}^{\kappa_u^T}(\mathbf{e'}_{u_j}^T)}_{\textit{interaction aggregation}}  \\
    & \oplus_{\kappa_u^T} \underbrace{\boldsymbol{M}_3 \otimes_{\kappa_u^T} \operatorname{map}_{\kappa_i^T}^{\kappa_u^T}(\mathbf{i'}_{u_j}^T)}_{\textit{item aggregation}}, 
    \label{user_aggregation}
\end{split}
\end{equation}

\begin{equation}
\begin{split}
    \mathbf{i}_{k}(T_n) \leftarrow \underbrace{\boldsymbol{M}_4 \otimes_{\kappa_i^T} \mathbf{h}_{i_k}(T_n)}_{\textit{item hidden aggregation}} 
    & \oplus_{\kappa_i^T} \underbrace{\boldsymbol{M}_5 \otimes_{\kappa_i^T} \exp_{\mathbf{o}}^{\kappa_i^T}(\mathbf{e'}_{i_k}^T)}_{\textit{interaction aggregation}}  \\
    & \oplus_{\kappa_i^T} \underbrace{\boldsymbol{M}_6 \otimes_{\kappa_i^T} \operatorname{map}_{\kappa_u^T}^{\kappa_i^T}(\mathbf{u'}_{i_k}^T)}_{\textit{user aggregation}}, \label{item_aggregation}
\end{split}
\end{equation}
where $\boldsymbol{M}$s are trainable matrices and $\mathbf{h}$ represents hidden state. The aggregation operator has three main components: hidden state aggregation, interaction aggregation and neighbor aggregation. 

Take the user aggregation shown in \cref{user_aggregation} and \cref{one step aggregation} as an example: the first component is the update of user $j$ hidden representation for several layers, which can be cascaded for multi-hop message aggregation, and the second component aggregates the associated edge information to user $j$ and the third component aggregates items information which is interacted by user $j$. \textit{It is worth noting that the problem of ignoring the information propagation among the same type of nodes can be solved by stacking several layers.}
$\mathbf{e'}_{u/i}^T$ in the interaction aggregation is either \textit{Early Fusion} $\mathbf{e'}_{u/i}^T = \textsc{Mlp}\left(\textsc{Pooling}\left(\mathbf{e} \in \mathcal{E}_{u/i}^{T}\right)\right)$ or \textit{Late Fusion} $\mathbf{e'}_{u/i}^T = \textsc{Pooling}\left(\textsc{Mlp}\left(\mathbf{e} \in \mathcal{E}_{u/i}^{T}\right)\right)$ and $\textsc{Pooling}(\cdot)$ can be one of the Mean Pooling and Max Pooling.
$\mathbf{i'}_{u}^T$ and $\mathbf{u'}_{i}^T$ in neighbor aggregation are the \textbf{weighted gyro-midpoints} \citep{ungar2010barycentric} (specified in the Appendix) in item and user stereographic spaces, respectively. Take $\mathbf{i'}_{u}^T$ as an example, which is defined as:
\begin{equation} \label{weighted midpoint of item}
    \begin{split}
        \mathbf{i'}_{u}^T = \textsc{Midpoint}_{\kappa_i^T} & \left(x_{j} \in \mathcal{N}_u^T\right) = \frac{1}{2} \otimes_{\kappa_i^T} \\
        & \left(\sum_{x_{j} \in \mathcal{N}_u^{T}} \frac{\alpha_{k} \lambda_{\mathbf{x}_{j}}^{\kappa}}{\sum_{x_{k} \in \mathcal{N}_u^{T}} \alpha_{k}\left(\lambda_{\mathbf{x}_{k}}^{\kappa}-1\right)} \mathbf{x}_{j}\right),
    \end{split}
\end{equation}
where $\alpha_k$ is attention weight and $\lambda_{\mathbf{x}}^{\kappa}$ is point-wise conformal factor.

After the aggregation of nodes and edges, we project embeddings to next-period manifolds for further training and predicting. The node updating equation is defined as follows:
\begin{equation} \label{user updating}
    \mathbf{u}_j(T_{n+1}) = \operatorname{map}_{\kappa_u^{T_n}}^{\kappa_u^{T_{n+1}}} \left(\mathbf{u}_j\left(T_n\right)\right), 
\end{equation}
\begin{equation} \label{item updating}
    \mathbf{i}_j(T_{n+1}) = \operatorname{map}_{\kappa_i^{T_n}}^{\kappa_i^{T_{n+1}}} \left(\mathbf{i}_j\left(T_n\right)\right),
\end{equation}
where the curvature $\kappa^{T_{n+1}}$ of next interval can be obtained from \cref{CurvNN}.

\subsection{Interaction Modeling}
We set the sequential interactions in Euclidean space instead of on Riemannian manifolds as user and item nodes.
The intuition behind such design is that the tangent space which is Euclidean is the common space of different Riemannian manifolds and users and items are linked by sequential interactions. 
Specifically, for an edge $e_k \in \mathcal{E}$, the integration of its corresponding attribute $\mathbf{X}_k \in \mathcal{X}$ and timestamp $t_k \in \mathcal{T}$ is regarded as the initialized embedding. In order to capture the temporal information of interactions, we formalize the time encoder as a function $\Phi_t: \mathbb{R} \rightarrow \mathbb{R}^d$ which maps a continuous time $t_k$ to a $d$-dimensional vector. In order to account for time translation invariance in consideration of the aforementioned batch setup, we adopt the \textit{harmonic encoder} \citep{Xu2020Inductive} as follows: 
\begin{equation} \label{time encoder}
    \begin{aligned}
        \phi(t) = \sqrt{\frac{1}{d}}\left[\cos\left(\omega_1 t + \theta_1\right); \cos\left(\omega_2 t + \theta_2\right)..;\cos\left(\omega_d t + \theta_d\right)\right], 
    \end{aligned}
\end{equation}
where $\omega$s and $\theta$s are learnable encoder parameters. and $\left[\cdot \,;\cdot\right]$ is the concatenation operator. \textit{The definition of \cref{time encoder} satisfies time translation invariance.} Thus, the initial embedding of the interaction $e_k$ can be represented as:
\begin{equation} \label{edge initial embedding}
    \begin{aligned}
        \mathbf{e}_k = \sigma(\boldsymbol{W_1} \cdot \left[\mathbf{X}_k \, ; \, \phi(t_k)\right]),
    \end{aligned}
\end{equation}
where $\sigma(\cdot)$ is the activation function and $\boldsymbol{W_1}$ is the training matrix. 

\subsection{Curvature Estimator}
Embedding vectors of the previous work are set on one fixed curvature manifold such as Poincar{\'{e}} Ball, while the fixed curvature is a strong premise assumption which cannot manifest the dynamics of the representation space. In fact, the curvature of the embedding space evolves over time, and thus we need to design curvature computation for both local curvature and global curvature, \ie, Ricci curvature and sectional curvature.


Sequential interaction networks are mostly bipartite graphs, which means there is no direct link connected between nodes of the same type. Therefore, to explore the local Ricci curvature in both user and item spaces, we need to extract new topologies from the input network. The rule of subgraph extraction is based on the shared neighbor relationship, \eg, users with common item neighbors are friends with each other. However, there will be several complete graphs which increase the complexity if we let all users with the same neighbors become friends. Here, we take sampling in extracting subgraph process to avoid long time cost. Then, the local curvature of any edge $(x, y)$ can be resolved by Ollivier-Ricci curvature $\kappa_{r}$ defined as follows:
\begin{equation} \label{ollivier ricci}
    \begin{aligned}
        \kappa_{r}(x, y)= 1 - \frac{W\left(m_{x}, m_{y}\right)}{d_{\mathcal{M}}^{\kappa}(x, y)},
    \end{aligned}
\end{equation}
where $W(\cdot, \cdot)$ is Wasserstein distance (or Earth Mover distance) which finds the optimal mass-preserving transportation plan between two probability measures. $m_{x}$ is the mass distribution of the node $x$ measuring the importance distribution of one node among its one-hop neighbors $\mathcal{N}(x)=\{x_1, x_2, ..., x_l\}$, which is defined as:
\begin{equation} \label{mass distribution}
    \begin{aligned}
        m_{x}^{\alpha}\left(x_{i}\right)= \begin{cases}\alpha & \text { if } x_{i}=x, \\ (1-\alpha) / l & \text { if } x_{i} \in \mathcal{N}(x), \\ 0 & \text { otherwise. }\end{cases}
    \end{aligned}
\end{equation}
where $\alpha$ is a hyper-parameter within $[0, 1]$. Similar to previous work \citep{Ye2020Curvature, sia2019ollivier}, we set $\alpha=0.5$. 

\IncMargin{1em}
\begin{algorithm}[ht]
    \caption{Curvature Observation}
    \label{curvature observation}
    \SetKwInOut{Input}{\textbf{Input}}
    \SetKwInOut{Output}{\textbf{Output}}
    \Input{An undirected graph $G$, iterations $n$}
    \Output{Observed curvature $\kappa_o$}
    \For{$m \in G$}{
        \For{$i = 1, ..., n$}{
           $b, c \in \mathop{Sample}(\mathcal{N}(m))$ and $a \in \mathop{Sample}(G)/\{m\}$\;
           Calculate $\gamma_{\mathcal{M}}(a, b, c)$\;
           Calculate $\gamma_{\mathcal{M}}(m;b, c;a)$\;
        }
        Let $\gamma(m) = \mathop{MEAN}(\gamma_{\mathcal{M}}(m;b, c;a))$\;
    }
    \textbf{Return:} $\kappa_o = \mathop{MEAN}(\gamma(m))$
\end{algorithm}

Let $\mathbf{R}$ represent the vector of Ricci curvatures. Inspired by the bilinear form between Ricci curvature and sectional curvature, we design the global sectional curvature neural network called \textit{CurvNN} as follows:
\begin{equation} \label{CurvNN}
    \begin{aligned}
        \kappa_e = \textsc{Mlp}\left(\mathbf{R}\right)^T \boldsymbol{W_2} \textsc{Mlp} \left(\mathbf{R}\right),
    \end{aligned}
\end{equation}
where $\boldsymbol{W_2}$ is the trainable matrix and $\textsc{Mlp}$ means multilayer perceptron. The bilinear design of \textit{CurvNN} allows us to obtain the sectional curvature of any sign.

The observation of the curvature is also important for us to train \textit{CurvNN}. Therefore, we adopt a variant of the Parallelogram Law to observe the sectional curvature \citep{gu2018learning, fu2021ace, bachmann20a}. Considering a geodesic triangle $abc$ on manifold $\mathcal{M}$ and $m$ is the midpoint of $bc$, two of their quantities are defined as: $\gamma_{\mathcal{M}}(a, b, c) := d_{\mathcal{M}}(a, m)^2 + d_{\mathcal{M}}(b, c)^2 /4 - \left(d_{\mathcal{M}}(a, b)^2 + d_{\mathcal{M}}(a, c)^2\right) / 2$ and $\gamma_{\mathcal{M}}(m;b, c;a) :=\frac{\gamma_{\mathcal{M}}(a, b, c)}{2d_{\mathcal{M}}(a, m)}$. The observed curvature is calculated in \cref{curvature observation}. The observed curvature $\kappa_o$ will supervise the training of \textit{CurvNN} to generate estimated curvature $\kappa_e$ accurately as possible.

\subsection{Training Objective}
Benefiting from the fact that the output of \ourmethod is embedded representation of users and items, we can treat them as input to various downstream tasks, such as interaction prediction. If we easily add up the distance from different spaces, the distortion will be huge due to different geometry metrics. Therefore, we consider our optimization from two symmetric aspects, \ie, user and item. Specifically, we first map the item embedding to user space and calculate the distance under user geometry metric, and vice versa as follows:
\begin{equation} \label{distance}
    \begin{split} 
        d_\mathbb{U}\left(\mathbf{u}, \mathbf{i}\right) = d_{\kappa_u}\left(\mathbf{u}, \operatorname{map}_{\kappa_i}^{\kappa_u}\left(\mathbf{i}\right)\right), \\
        d_\mathbb{I}\left(\mathbf{u}, \mathbf{i}\right) = d_{\kappa_i}\left(\operatorname{map}_{\kappa_u}^{\kappa_i}\left(\mathbf{u}\right), \mathbf{i}\right).
    \end{split}
\end{equation}

Then, we regard Fermi-Dirac decoder, a generalization of Sigmoid, as the likelihood function to compute the connection probability between users and items in both spaces. Since we convert the distance into probabilities, the probabilities then can be easily summed. Moreover, we are curious about which geometric manifolds are more critical for interaction prediction, and then assign more weight to the corresponding probability. The likelihood function and probability assembling are defined as follows:
\begin{align}
    P_{\mathbb{U}/\mathbb{I}}(e | \Theta) 
    & =\frac{1}{\exp\left(\left(d_{\mathbb{U}/\mathbb{I}}(\mathbf{u}, \mathbf{i})-r\right) / t\right)+1},  \label{probability} \\
    P(e | \Theta) 
    & = \rho P_{\mathbb{U}} + (1-\rho) P_{\mathbb{I}}, \label{probability sum}
\end{align}
where $\Theta$ represents all learnable parameters in \ourmethod, $r, t>0$ are hyperparameters of Fermi-Dirac decoder, and $\rho$ is the learnable weight. Users and items share the same formulation of the likelihood.

The main optimization is based on the idea of Pull-and-Push. We hope that the embedding in batch $T_n$ is capable of predicting the interaction in batch $T_{n+1}$, thus we utilize the graph in batch $T_{n+1}$ to supervise the previous batch. In particular, we expect a greater probability for all interactions that actually occur in next batch, but a smaller probability for all the negative samples. Given the large number of negative samples, we determine to use negative sampling (Gaussian) with rate of 20\% and it can also avoid the problem of over-fitting. For the curvature estimator component, we wish that \textit{CurvNN} can learn the global sectional curvature from the local structure (Ricci curvature) without observing in the prediction task. Therefore, we make the Euclidean norm of estimated curvature $\kappa_e$ and observed curvature $\kappa_o$ as our optimizing target. 
We calculate the total loss as follows:
\begin{equation} \label{optim target}
    \begin{split}
        \Theta^{*} = \mathop{argmax}\limits_{\Theta} \sum_{e \in \mathcal{E}} & \ln [P(e |\Theta)] \\
        - & \mathop{Sampling} \left(\sum_{e \notin \mathcal{E}} \ln [P(e|\Theta)]\right) + \Vert \kappa_{e} - \kappa_{o} \Vert_2.
    \end{split}
\end{equation}

\subsection{Computational Complexity Analysis}
The procedure of training \ourmethod is given in appendix. The curvature estimator and cross-geometry operations are two main processes of \ourmethod. To avoid excessive parameters, we use a shared curvature estimator for both user and item spaces. The primary computational complexity of \ourmethod is approximated as  $O(|\mathcal{U}|DD'+|\mathcal{I}|DD'+|\mathcal{E}|D'^2)+O(V^3\log V)$, where $|\mathcal{U}|$ and $|\mathcal{I}|$ are number of user and item nodes and $|\mathcal{E}|$ is the number of interactions. $D$ and $D'$ are the dimensions of input and hidden features respectively and $V$ is the total number of vertices in the Wasserstein distance sub-problem. The former $O(\cdot)$ is the computational complexity of all aggregating and updating operations and the computation of other parts in our model can be parallelized and is computationally efficient. The latter $O(\cdot)$ is the complexity of Olliver-Ricci curvature. In order to avoid repeated calculation of Ricci curvature, we adopt an \textbf{OFFLINE} pre-computation that calculates Ricci curvatures according to different batch sizes and saves it in advance since the subgraph in each batch does not change.

\section{Experiments} \label{Experiments}
\subsection{Datasets and Compared Algorithms}
\textbf{Datasets.}
 We conduct experiments on five dynamic real-world datasets: \textbf{MOOC}, \textbf{Wikipedia}, \textbf{Reddit}, \textbf{LastFM} and \textbf{Movielen} to evaluate the effectiveness of \ourmethod. The description and statistic details of these datasets are shown in Appendix.



\noindent \textbf{Compared Algorithms.} 
To demonstrate the effectiveness of our method, we compare our model with nine state-of-the-art models, categorized as follows: 
\begin{itemize}
    \item \textbf{Recurrent sequence models:} we compare with different flavors of RNNs trained for item-sequence data: LSTM, Time-LSTM \citep{time-lstm}, RRN \citep{RRN}.
    \item \textbf{Random walk based models:} CAW \citep{CAW}, CTDNE \citep{CTDNE} are two temporal network models adopting causal and anonymous random walks.
    \item \textbf{Sequence network embedding models:} in this category, JODIE \citep{jodie}, HILI \citep{hili} and DeePRed \citep{deepred} are three state-of-the-art methods in generating embeddings from sequential networks employing recursive networks.
    \item \textbf{Hyperbolic models:} we compare \ourmethod with HGCF \citep{HGCF} which is a hyperbolic method on collaborative filtering.
\end{itemize}

\begin{table*}[ht]
    \caption{Future interaction prediction: Performance comparison in terms of mean reciprocal rank (MRR) and Recall@10. The best results are in \textbf{bold} and second best results are \underline{underlined}.}
    \label{tab:main-results-with-onehot}
    \centering
    \resizebox{\textwidth}{!}{
    \begin{tabular}{l|cc|cc|cc|cc|cc}
        \toprule
        Datasets $\rightarrow$&\multicolumn{2}{c|}{MOOC}&\multicolumn{2}{c|}{Wikipedia}&\multicolumn{2}{c|}{Reddit}&\multicolumn{2}{c|}{LastFM}&\multicolumn{2}{c}{Movielen} \\
        Methods $\downarrow$&MRR&Recall@10&MRR&Recall@10&MRR&Recall@10&MRR&Recall@10&MRR&Recall@10 \\
        \midrule
        \midrule
        LSTM & 0.055 & 0.109 & 0.329 & 0.455 & 0.355 & 0.551 & 0.062 & 0.119 & 0.031 & 0.060 \\
        Time-LSTM \citep{time-lstm} & 0.079 & 0.161 & 0.247 & 0.342 & 0.387 & 0.573 & 0.068 & 0.137 & 0.046 & 0.084 \\
        RRN \citep{RRN} & 0.127 & 0.230 & 0.522 & 0.617 & 0.603 & 0.747 & 0.089 & 0.182 & 0.072 & 0.181\\
        CAW \citep{CAW} & 0.200 & 0.427 & 0.656 & 0.862 & 0.672 & 0.794 & 0.121 & 0.272 & 0.096 & 0.243 \\
        CTDNE \citep{CTDNE} & 0.173 & 0.368 & 0.035 & 0.056 & 0.165 & 0.257 & 0.01 & 0.01 & 0.033 & 0.051 \\
        JODIE \citep{jodie} & \underline{0.465} & 0.765 & 0.746 & 0.822 & 0.726 & 0.852 & 0.195 & 0.307 & 0.428 & 0.685 \\
        HILI \citep{hili} & 0.436 & \underline{0.826} & 0.761 & 0.853 & 0.735 & 0.868 & 0.252 & \underline{0.427} & \underline{0.469} & \underline{0.784} \\
        DeePRed \citep{deepred} & 0.458 & 0.532 & \textbf{0.885} & \textbf{0.889} & \textbf{0.828} & \underline{0.833} & \underline{0.393} & 0.416 & 0.441 & 0.472 \\
        HGCF \citep{HGCF} & 0.284 & 0.618 & 0.123 & 0.344 & 0.239 & 0.483 & 0.040 & 0.083 & 0.108 & 0.260 \\
        \hline
        \hline
        \ourmethod(ours) & \textbf{0.586} & \textbf{0.885} & \underline{0.793} & \underline{0.865} & \underline{0.825} & \textbf{0.883} & \textbf{0.425} & \textbf{0.466} & \textbf{0.511} & \textbf{0.819} \\
        \bottomrule
    \end{tabular}}
\end{table*}

\subsection{Future Interaction Prediction} \label{future interaction prediction}
The goal of the future interaction prediction task is to predict the next item a user is likely to interact with based on observations of recent interactions. 
We here use two metrics, mean reciprocal rank (MRR) and Recall@10 to measure the performance of all methods. The higher the metrics, the better the performance. 
For fair comparison, all models are run for 50 epochs and the dimension of embeddings for most models are set to 128. We use a chronological train-valid-test split with a ratio of 80\%-10\%-10\% following other baselines. We use the validation set to tune \ourmethod hyperparameters using Adam optimization with a learning rate of $3 \times 10^{-3}$.

\begin{figure}[ht]
    \centering
    \includegraphics[width=\linewidth]{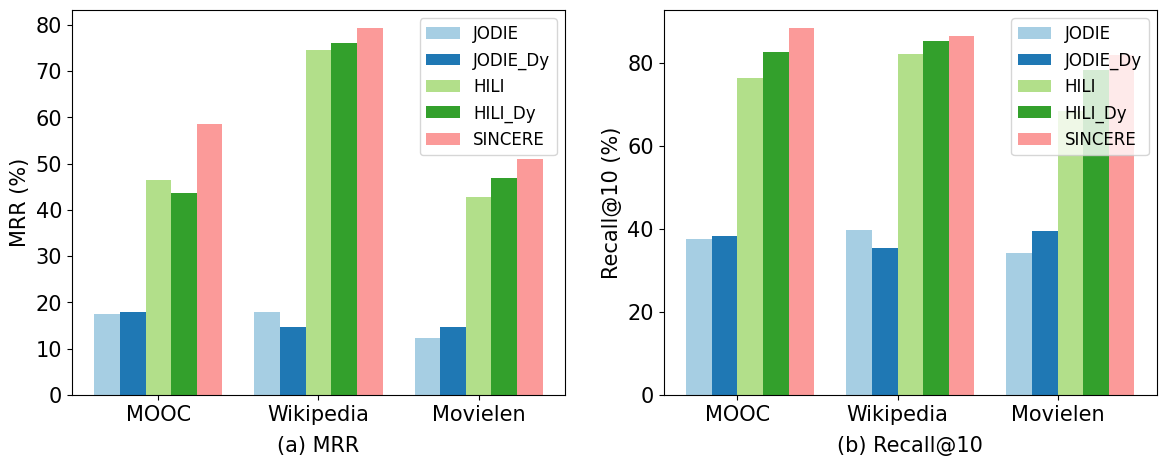}
    \caption{Comparison between methods and the ones without static embeddings on future interaction prediction.}
    \label{MRR&REC comparison}
    \Description{Comparison between methods and the ones without static embeddings on future interaction prediction.}
\end{figure}
During testing, given a former period of interactions, \ourmethod first estimates the sectional curvature of the next period based on the structure information, then calculates the link probability between the specific user and all items and predicts a ranked top-$k$ items list. 
The experimental results are reported in \cref{tab:main-results-with-onehot}, where all the baselines use the parameters of the best performance according to the original papers. In general, our method performs the best against other methods, demonstrating the benefits of dual spaces design and representation spaces dynamics. Among all methods, we find that \ourmethod and DeePRed have the smallest gaps between MRR and Recall@10, which means not only can these two models find the exact items the user may interact with, but also can distinguish the ground truth from negative samples with higher ranks. 

While predicting future interactions, we discover that the performance of two state-of-the-art methods JODIE and HILI have little vibration with the embedding size from 32 to 256 compared with other methods, which means they are insensitive to the embedding size. We argue that the robustness to embedding size is actually irrelevant to the merit of the model itself and we argue that JODIE and HILI have a relatively strong reliance on static embeddings which are one-hot vectors. To verify this, we have further investigation on the results of removing static embeddings (one-hot vectors). In \cref{MRR&REC comparison}, JODIE\_Dy and HILI\_Dy indicates the ones with no static embedding, JODIEs are in blue schemes and HILIs in green. The experimental result shows that there is a huge gap once the one-hot vectors are removed which verifies our assumption, while \ourmethod performs the best in terms of MRR and Recall@10. Besides, HILI relies less on the static embeddings for the greater gain between HILI and HILI\_Dy compared to JODIE.  
\begin{table*}[ht]
    \caption{Comparison on different variants of \ourmethod.}
    \label{tab:ablation-results}
    \centering
    \resizebox{0.85\textwidth}{!}{
    \begin{tabular}{l|cc|cc|cc|cc}
        \toprule
        Datasets $\rightarrow$&\multicolumn{2}{c|}{MOOC}&\multicolumn{2}{c|}{Wikipedia}&\multicolumn{2}{c|}{Reddit}&\multicolumn{2}{c}{Movielen} \\
        Methods $\downarrow$&MRR&Recall@10&MRR&Recall@10&MRR&Recall@10&MRR&Recall@10 \\
        \midrule
        \midrule
        \ourmethod&\textbf{0.586}&\textbf{0.885}&\textbf{0.793}&\textbf{0.865}&\textbf{0.825}&\textbf{0.883}&\textbf{0.511}&\textbf{0.819} \\
        \ourmethod-Static&0.469&0.813&0.747&0.834&0.703&0.775&0.402&0.664 \\
        \ourmethod-$\mathbb{E}$&0.397&0.674&0.692&0.758&0.682&0.723&0.374&0.616 \\
        \bottomrule
    \end{tabular}}
\end{table*}

\subsection{Ablation Analysis}
The introduction of co-evolving $\kappa$-stereographic spaces is the basis of \ourmethod. In order to analyze the effect of dynamic modeling and co-evolving $\kappa$-stereographic spaces, we conduct ablation experiments on four datasets with two variant models named \ourmethod-Static and \ourmethod-$\mathbb{E}$, respectively. 
\begin{figure}[ht]
    \centering
    \includegraphics[width=0.95\linewidth]{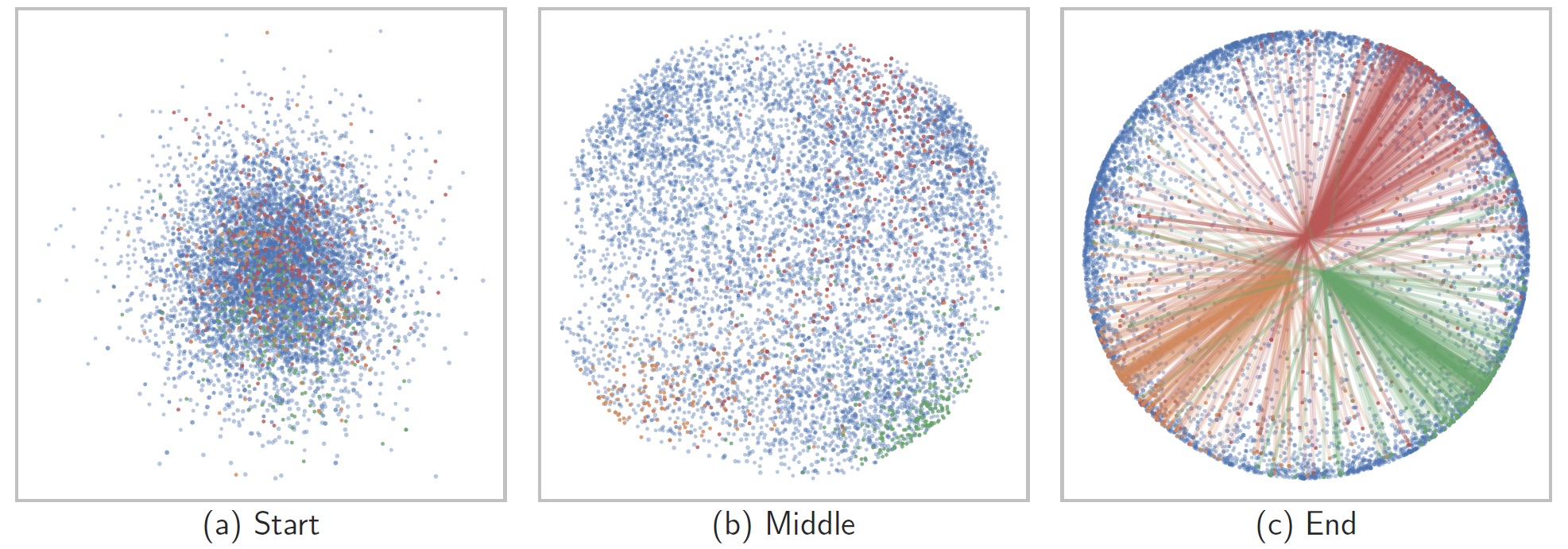}
    \caption{A visualization of three different status of \ourmethod-Static on Movielen dataset. We follow and color the movie nodes viewed by three users in \textcolor{expred}{red}, \textcolor{exporange}{orange} and \textcolor{expgreen}{green}. Others are in \textcolor{expblue}{blue}. Central convergence points are the three users we pick and map from user space. }
    \label{fig:experiment visualization}
    \Description{A visualization of three different status of SINCERE-Static on Movielen dataset.}
\end{figure}

For \ourmethod-Static, we setup the static curvature from the training beginning the same as last-batch curvature of the previous experiment to make the spaces steady. For \ourmethod-$\mathbb{E}$, we fix the curvature to zero as the space is Euclidean. 
As shown in \cref{tab:ablation-results}, the performance of \ourmethod-Static is closer to \ourmethod compared with \ourmethod-$\mathbb{E}$ which is because the curvature set in \ourmethod-Static can be seen as the pseudo-optimal curvature. The small gap indicates that there is some information loss in the training process even though the pseudo-optimal curvature is given. \cref{fig:experiment visualization} gives an visualization of \ourmethod-Static on Movielen dataset with curvature set to $-1$. The performance of fixing the space as Euclidean is somehow less competitive, which confirms that introducing dual dynamic Riemannian manifolds is necessary and effective. 

\subsection{Hyperparameter Sensitivity Analysis} \label{hyperparameter sensitivity analysis}
To further investigate the performance of \ourmethod, we conduct several experiments for \ourmethod with different setups to analyze its sensitivity to hyperparameters. 

\begin{figure}[ht]
    \centering
    \includegraphics[width=\linewidth]{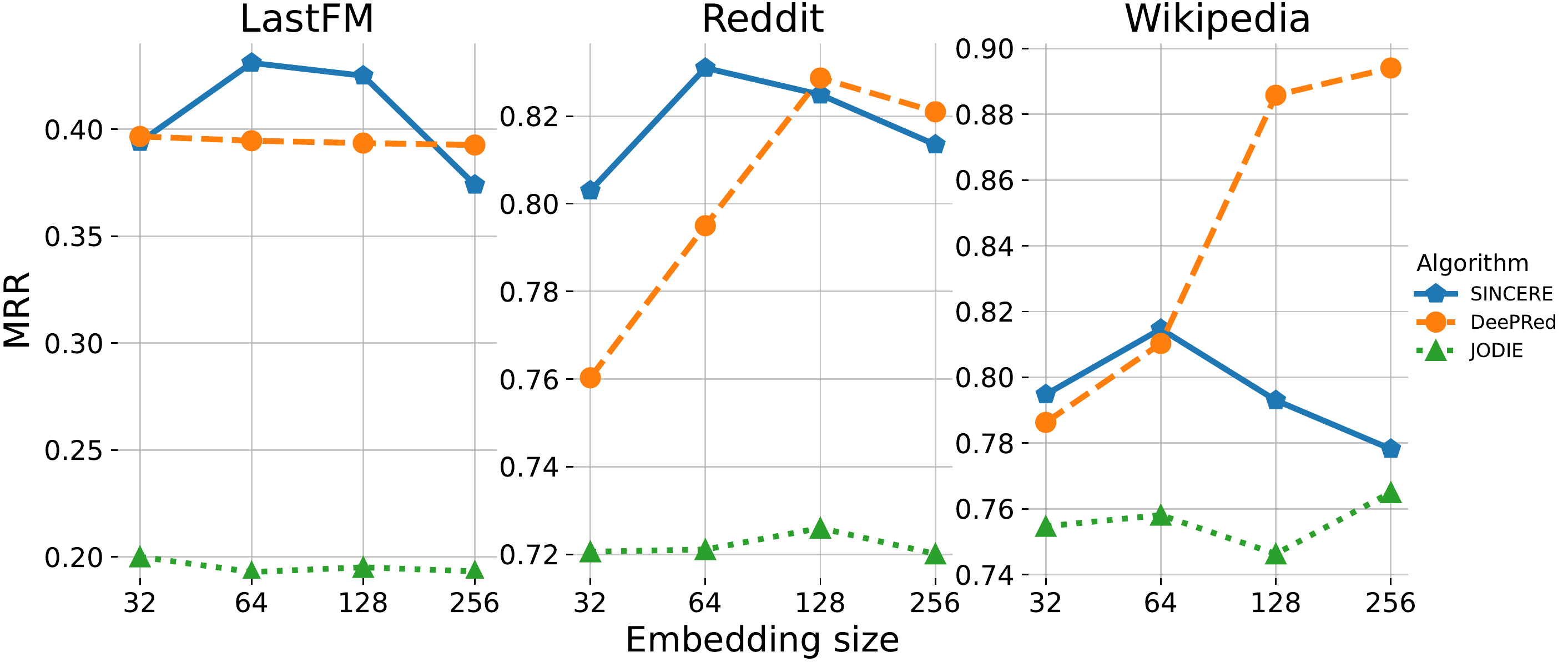}
    \caption{Effect of embedding size.}
    \label{fig:embedding_size}
    \Description{Effect of embedding size.}
    \vspace{-2mm}
\end{figure}
\cref{fig:embedding_size} shows the impact of embedding dimension on the performance of three methods. $128$ seems an optimal value for DeePRed in most cases. As we discuss in \cref{future interaction prediction}, JODIE counts on the static embeddings which makes embedding size have almost no influence on it. However, we find that in some cases such as LastFM and Wikipedia, \ourmethod performs better with dimension $64$ than $128$ and we argue that it is due to the strong representational power of hyperbolic spaces. In these cases, $64$ is capable enough to represent the networks.

\begin{figure}[ht]
    \centering
    \includegraphics[width=\linewidth]{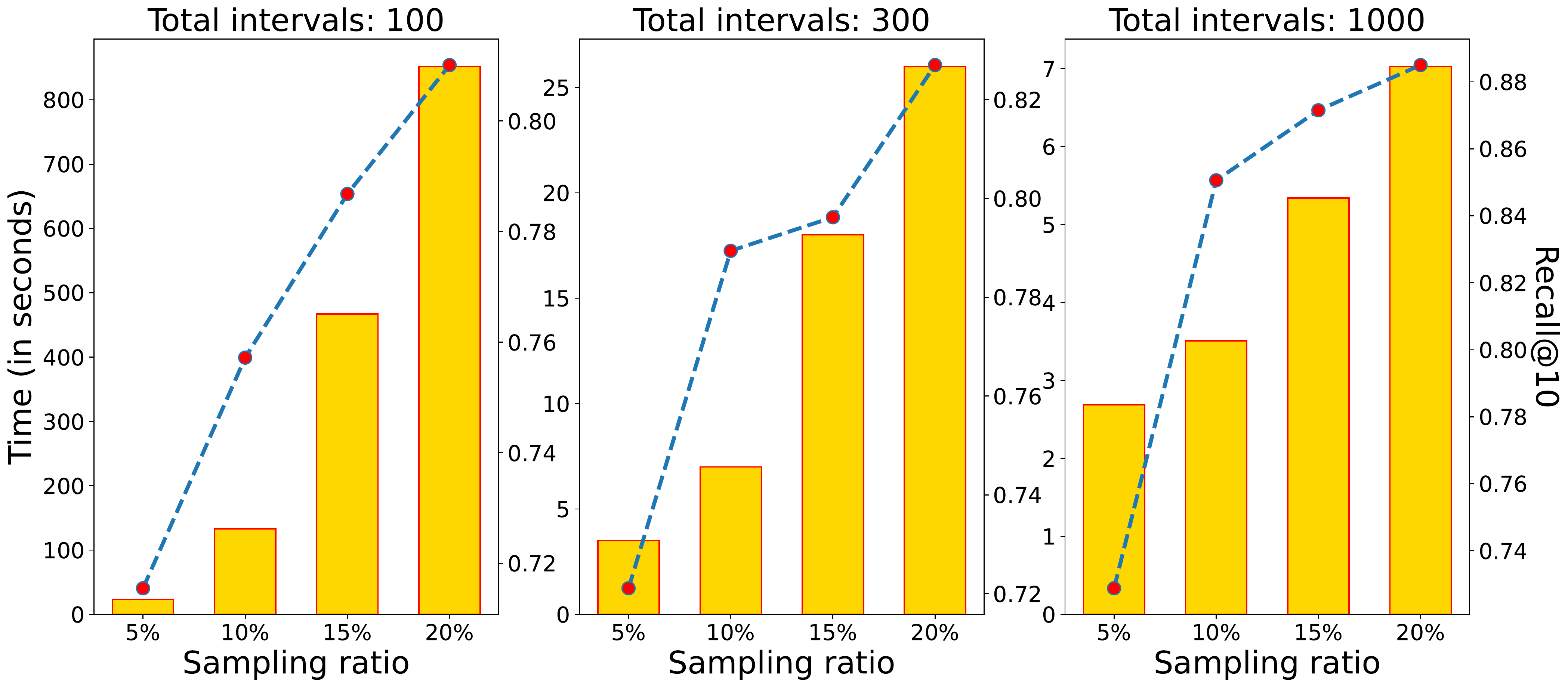}
     \vspace{-0.15in}
    \caption{Effect of sample ratio on MOOC dataset.}
    \label{fig:sample_rate}
    \Description{Effect of sample ratio on MOOC dataset.}
    \vspace{-0.15in}
\end{figure}
The time spent in computing Ricci curvatures and the influence of subgraph extraction sampling ratio on \ourmethod performance are shown in \cref{fig:sample_rate}. We observe that the computing time decreases sharply with the number of total intervals increase. Besides, the performance (Recall@10) levels off from the sampling ratio $15\%$ to 20\% as interval number grows. Such finding informs us that instead of increasing sampling rate to improve performance, it is far better to set a higher interval number considering the computational complexity of Ricci curvatures.

\vspace{-0.2in}

\section{Related Work}
\noindent \textbf{Sequential Graph Representation.}
Graph representation assigns individual node an embedding integrating structure and attribute information. Sequential interaction graph representation takes temporal dependency into consideration \citep{kazemi2020representation, aggarwal2014evolutionary}, and temporal graph neural networks achieve great success recently \citep{Xu2020Inductive, wang2021MeTA, zuo2018embedding, gupta2022tigger, sun2022selfcikm}. Among them, RNN based models \citep{time-lstm, RRN, tlstm, latentcross} are the most popular due to effectiveness on sequential data. CTDNE \citep{CTDNE} and CAW \citep{CAW} adopt random walk to temporal graph modeling. Co-evolution models consider the influence between users and items and achieve great success \citep{deepcoevolve, jodie, hili, deepred}. For systematic and comprehensive reviews, readers may refer to \citep{kazemi2020representation, aggarwal2014evolutionary}.

\noindent \textbf{Riemannian Graph Learning.}
 Hyperbolic and spherical manifolds are typical instances of Riemannian manifolds and recently serve as the alternative of Euclidean spaces, solving distortion problems.
A series of Riemannian graph models are proposed \citep{liu2019hyperbolic, brehmer2020flows, sonthalia2020tree, chami2020from, sun2021hyperbolic, sun2022selfContinual}. Recently, to match the wide variety of graph structures, the study of mixed-curvature space has been conducted. For instance, MVAE \citep{skopek2020mixed-curvature} and SelfMGNN \citep{sun2022self} are two recent representatives to capture the complex graph structure in product manifolds. As another line of work, GIL \citep{zhu2020graph} attempts to embed a graph into the Euclidean space and hyperbolic space simultaneously for interaction learning. 

\section{Conclusion}
In this paper, we take the first attempt to study the sequential interaction network representation learning in co-evolving Riemannian spaces, and present a novel model \ourmethod. Specifically, we first introduce a couple of co-evolving Riemannian representation spaces for both types of nodes bridged by the Euclidean-like tangent space. Then, we design the sectional curvature estimator and geometry-cross methods for space evolution trend prediction and message propagation. Finally, we conduct extensive experiments on several real-world datasets and show the superiority of \ourmethod. 

\clearpage
\bibliographystyle{ACM-Reference-Format}
\balance
\begin{acks}
This work was supported in part by National Natural Science Foundation of China under Grant U1936103, 61921003 and 62202164.
\end{acks}
\bibliography{custom}

\end{document}